\documentclass[default,iicol]{sn-jnl}
\usepackage{float}

\usepackage{textcomp}
\usepackage{stfloats}
\usepackage{url}
\usepackage{verbatim}
\usepackage{graphicx}
\usepackage{multirow}
\usepackage{textcomp}
\usepackage{booktabs} 
\usepackage{amssymb}
\usepackage{pifont}
\definecolor{SeaGreen4}{RGB}{0,205,102} 
\definecolor{SlateBlue}{RGB}{106,90,205} 
\definecolor{DarkRed}{RGB}{178,34,34}
\renewcommand{\cmark}{\ding{51}}%
\renewcommand{\xmark}{\ding{55}}%

\begin{document}

\title[Article Title]{Treat Stillness with Movement: Remote Sensing Change Detection via Coarse-grained Temporal Foregrounds Mining}  


\author[1]{\fnm{Xixi} \sur{Wang$^{\dag}$}}
\author[1]{\fnm{Zitian} \sur{Wang$^{\dag}$}}
\author[1]{\fnm{Jingtao} \sur{Jiang}}
\author[2]{\fnm{Lan} \sur{Chen}$^{(\textrm{\Letter})}$}
\author[1]{\fnm{Xiao} \sur{Wang}}
\author[1]{\fnm{Bo} \sur{Jiang}}

\affil[1]{\orgdiv{School of Computer Science and Technology}, \orgname{Anhui University}, \orgaddress{\city{Hefei} \postcode{230601},  \country{China}}} 
\affil[2]{\orgdiv{School of Electronic and Information Engineering}, \orgname{Anhui University}, \orgaddress{\city{Hefei} \postcode{230601},  \country{China}}}


\abstract{
Current works focus on addressing the remote sensing change detection task using bi-temporal images. Although good performance can be achieved, however, seldom of they consider the motion cues which may also be vital. In this work, we revisit the widely adopted bi-temporal images-based framework and propose a novel Coarse-grained Temporal Mining Augmented (CTMA) framework. To be specific, given the bi-temporal images, we first transform them into a video using interpolation operations. 
Then, a set of temporal encoders is adopted to extract the motion features from the obtained video for coarse-grained changed region prediction. 
Subsequently, we design a novel Coarse-grained Foregrounds Augmented Spatial Encoder module to integrate both global and local information.
We also introduce a motion augmented strategy that leverages motion cues as an additional output to aggregate with the spatial features for improved results. 
Meanwhile, we feed the input image pairs into the ResNet to get the different features and also the spatial blocks for fine-grained feature learning. 
More importantly, we propose a mask augmented strategy that utilizes coarse-grained changed regions, incorporating them into the decoder blocks to enhance the final changed prediction. 
Extensive experiments conducted on multiple benchmark datasets fully validated the effectiveness of our proposed framework for remote sensing image change detection. 
The source code of this paper will be released on \url{https://github.com/Event-AHU/CTM_Remote_Sensing_Change_Detection}. 
}





\maketitle

\section{Introduction} \label{sec1}

Remote sensing image change detection targets finding the variable pixel-level regions between given two images. This task can be used in many practical scenarios, including damage assessment, urban studies, ecosystem monitoring, agricultural surveying, and resource management. Although good performance can already be achieved in some simple scenarios, remote sensing change detection is still a challenging task in extreme cases.

\begin{figure*}
    \centering
    \includegraphics[width=1\textwidth]{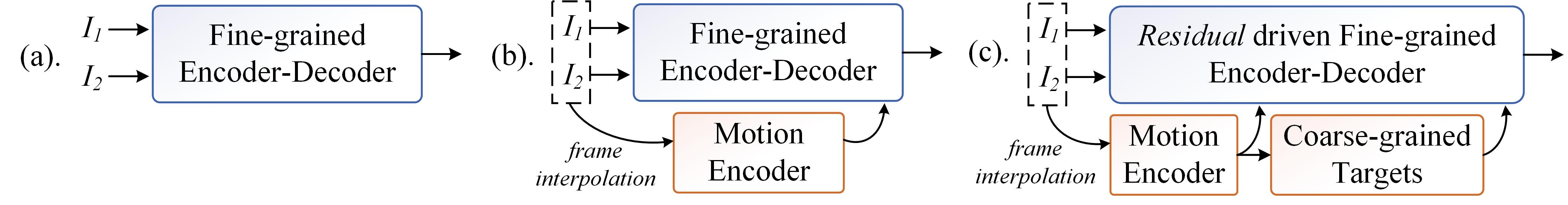}
    \caption{Comparison between existing 
        (a). Fine-grained encoder-decoder framework for RSCD; 
        (b). Motion-augmented fine-grained encoder-decoder framework for RSCD; 
        (c). Our newly proposed coarse-grained temporal foregrounds mining for RSCD. 
        Note that, the $I_1$ and $I_2$ are input image pairs. 
        }
    \label{fig:firstIMG}
\end{figure*}

Existing researchers usually adopt Convolutional Neural Networks (CNN)~\cite{fu2020camera} and Transformers~\cite{liu2024vision} to build their backbones for remote sensing image change detection, as illustrated in Fig.~\ref{fig:firstIMG}(a). 
Specifically, 
Chen et al.~\cite{chen2021remote} propose a Bitemporal Image Transformer (BIT) that mines the contexts within the spatial-temporal domain effectively. 
The Visual change Transformer (VcT) proposed by Jiang et al.~\cite{jiang2023vct} finds that the mining of the common background information helps the consistent representations which further enhances the visual change detection task. 
Tang et al.~\cite{tang2023fredecRSCD} propose the frequency decoupling interaction (FDINet) for the object fine-grained change detection task. 
In addition to the fine-grained encoder-decoder framework proposed for remote sensing image change detection, some researchers also exploit the motion features to further augment the final performance, as shown in Fig.~\ref{fig:firstIMG}(b). 
For example, 
Gan et al.~\cite{gan2024RFLCDNet} propose the RFL-CDNet framework and achieve high-performance change detection by richer feature learning. 
Lin et al.~\cite{lin2022transition} propose the pair-to-video change detection (P2V-CD) framework to address the issues of 
incomplete temporal modeling, and space-time coupling, which further improves the change detection results compared with existing ones.

Although these models achieve good performance on current benchmark datasets, however, we think their performance may still be limited by the following issues: 
1). Existing change detection models mainly adopt a single-stage framework, focusing on how to extract and relate the feature representations from a given pair of images. However, few algorithms consider incorporating temporal information to enhance change detection. 
2). Current works focus on extracting multi-scale spatial features for changed region prediction. However, seldom of they consider the residual difference between the given input pairs. In addition, they pay less attention to the coarse-grained target regions which may be also useful for final changed region prediction. 
Therefore, it is natural to raise the following question: \textit{how can we design a novel remote sensing change detection framework that effectively leverages motion cues, residual differences, and multi-scale information?}

To address the aforementioned issues, in this paper, we propose a novel Coarse-grained Temporal Mining Augmented framework for remote sensing image change detection task that mines the coarse-grained foregrounds to augment the fine-grained spatial encoder network, as shown in Fig.~\ref{fig:firstIMG} (c). 
The key insight of this paper is that the motion cues are very important for the change detection task, but the dense video-level motion information extraction is always ignored or not given sufficient attention by previous researchers. To be specific, as shown in Fig.~\ref{framework}, given the image pairs, we first transform them into dense frames using interpolation. Then, we propose to extract the motion features using the temporal encoder network and the possible changed masks can be obtained using a convolutional layer. 
The binary target mask is used to extract the pixel regions by multiplying them with the input image pairs. Meanwhile, we also design a Coarse-grained Foregrounds Augmented Spatial Encoder module that considers motion features, residual differences, and multi-scale features for accurate change detection. Note that, the residual difference is obtained by subtracting the feature maps using another one and will be injected into the decoding networks. Finally, the predictions of the decoder network and the segmented targets (via mask augmented strategy) are added for the final prediction.

To sum up, the contributions of this paper can be summarized as the following two main aspects:

$\bullet$ We propose a novel Coarse-grained Temporal Mining Augmented framework that first mines the coarse-grained foregrounds for remote sensing image change detection, and then augments the fine-grained spatial encoder network. It adaptively fuses the spatial, temporal, and multi-scale features for accurate detection. 

$\bullet$ We design a novel Coarse-grained Foregrounds Augmented Spatial Encoder module, which effectively integrates both global and local information. By introducing motion augmented and mask augmented strategies, this module obtains more precise change detection. 

$\bullet$ We conducted extensive experiments on multiple remote sensing change detection datasets and the experimental results fully validated the effectiveness of our proposed framework.

\textit{The following of this paper is organized as}: 
We give an introduction to related works on remote sensing change detection and spatial-temporal feature learning in Section~\ref{relatedworks}. Then, in Section~\ref{sec:methods}, we will describe the main frameworks proposed in this paper, with a focus on the temporal encoder network, coarse-grained foregrounds augmented spatial encoder and loss functions. After that, in Section~\ref{experiments}, we will introduce the experimental settings, including datasets, evaluation metrics, implementation details, and also the results on benchmark datasets by comparing them with other state-of-the-art models, ablation studies, and related visualizations. Finally, in Section~\ref{conclusion}, we conclude this paper and propose possible research directions as our future works.

\section{Related Work} \label{relatedworks}

In this section, we provide a brief introduction to the works focused on Remote Sensing Image Change Detection~\cite{wu2020computational}  and Spatial-Temporal Feature Learning.

\subsection{Remote Sensing Image Change Detection} 
Remote sensing image change detection is an important application field in remote sensing technology, and its purpose is to detect changes in surface features by comparing remote sensing images from different periods. Most of the existing methods are based on spatial modeling of two images, and direct pairwise difference or splicing of images for change detection.
Jiang et al.~\cite{jiang2023vct} propose that VcT enables the model to learn a consistent representation between two images by mining spatial contextual information. Goswami et al.~\cite{electronics11030431} used decision tree algorithms and post-classification comparisons of separation matrices, as well as image differencing in algebraic techniques to detect two images. Li et al.~\cite{10034814} propose a novel lightweight network, called A2Net, that recognizes changes by moving network-extracted features in combination with progressive feature aggregation and supervised attention. Zhou et al.~\cite{10123995} proposed a Context Aggregation Network (CANet) for mining cross-image contextual information between all training images to further enhance contextual representation within a single image.
Since the development phase of change detection, a great deal of work has been devoted to enhancing the representation of spatial features. However, specialized modeling of time has long been neglected~\cite{lin2022transition}. Methods focusing on the time dimension include two main categories, recurrent neural network (RNN) ~\cite{medsker2001recurrent} based methods and attention-based ~\cite{bahdanau2014neural} methods. 

RNNs process sequential data with a memory function that captures the backward and forward relationships in a sequence, characterized by accepting inputs and a "hidden state" and outputting a new hidden state at each time step. This design allows information to be passed between time steps, thus capturing long-term dependencies in the sequence, and some improved structures have been developed, such as the Long Short-Term Memory (LSTM) network. Mou et al.~\cite{mou2018learning}presents the first recursive convolutional network architecture for multi-temporal remote sensing image analysis, utilizing a novel recursive convolutional neural network (ReCNN) architecture for change detection in multi-spectral images. The network combines a convolutional neural network (CNN)~\cite{jia2021performance} and RNN to learn joint spectral-spatial-temporal feature representations in a unified framework. 
Chen et al~\cite{chen2019change} proposed a deep twinned convolutional multilayer recurrent neural network (SiamCRNN) for change detection in multi-temporal VHR images, where a multilayer recurrent neural network stacked with Long Short-Term Memory (LSTM) units is responsible for mapping the extracted spatial-spectral features to new latent feature space and mining the change information in it. 

The attention mechanism is a technique used to improve the performance of deep learning models, especially when processing sequential data, such as in tasks like natural language processing and image processing. The core idea of Attention Mechanism is to allow the model to selectively focus on certain important parts of the input data when processing, instead of treating all the information equally. Some researchers have begun to use attention-based methods to model temporal relationships in change detection tasks. Chen et al.~\cite{chen2020spatial} designed a change detection self-attention mechanism to model spatial-temporal~\cite{chen2024deep} relations, integrating a new self-attention module in the feature extraction process. Wang et al.~\cite{10530342} proposed a two-temporal-phase attention-sharing network that fully takes into account the spatial location and scale variations of different objects.

The methods described above are able to adaptively learn the temporal dependency between two temporal images. However, the input sequence consists of only two images. This suffers from incomplete temporal modeling and spatial-temporal coupling, for which Lin et al. ~\cite{lin2022transition} proposes a more explicit and sophisticated temporal modeling approach and constructs a pseudo-transitional video that carries rich temporal information to interpret change detection as a video comprehension problem.

\subsection{Spatial-Temporal Feature Learning} 
Spatial-Temporal feature learning~\cite{wang2024STsideTuningPAR, wang2024eventVOT, wang2020deepSTDMSurvey} is a technique for extracting and modeling features of data that vary in spatial and temporal dimensions. It has a wide range of applications in many fields, including computer vision, neuroscience, traffic prediction, and climate research. Spatial-temporal feature learning captures the dynamic behavior of data at different locations and points in time, providing a more comprehensive understanding and analysis. Aghili et al.~\cite{aghili2023spatial} proposed Spatial-Temporal Linear Feature Learning (STLFL), which is an improved linear discriminant analysis technique focused on extracting high-level features of P300 event-related brain potentials. Tan et al.~\cite{tan2023learning} proposed a new dynamic spatial-temporal graph data modeling framework for constructing spatial-temporal adjacency graphs through the lens of graph products. Lu et al.~\cite{lu2023learning} proposed the Spatial-Temporal Fusion (STF) module to learn implicit neural representations from spatial-temporal coordinates and features queried from RGB frames and events. Cui et al.~\cite{cui2023multi} proposed a Transformer-based gait recognition framework that introduces a spatial fusion module (SFM) and a temporal fusion module (TFM) for efficiently fusing spatial and temporal level feature information, respectively. Wang et al.~\cite{wang2024hardvs} propose an ESTF framework for event-based action recognition based on spatial and temporal Transformer networks. 
Inspired by these works, in this paper, we also exploit the spatial-temporal features to further augment the performance of changed region detection. 

\section{The Proposed Method}\label{sec:methods}    

In this section, we introduce a novel Coarse-grained Temporal Mining Augmented (CTMA) framework for remote sensing change detection task. As illustrated in Fig.~\ref{framework}, it mainly consists of Temporal Encoder (TE) and Coarse-grained Foregrounds Augmented Spatial Encoder (CFA-SE). The details will be elaborated below.

\subsection{Overview} 
\begin{figure*} 
\centering
\includegraphics[width=0.8\textwidth]{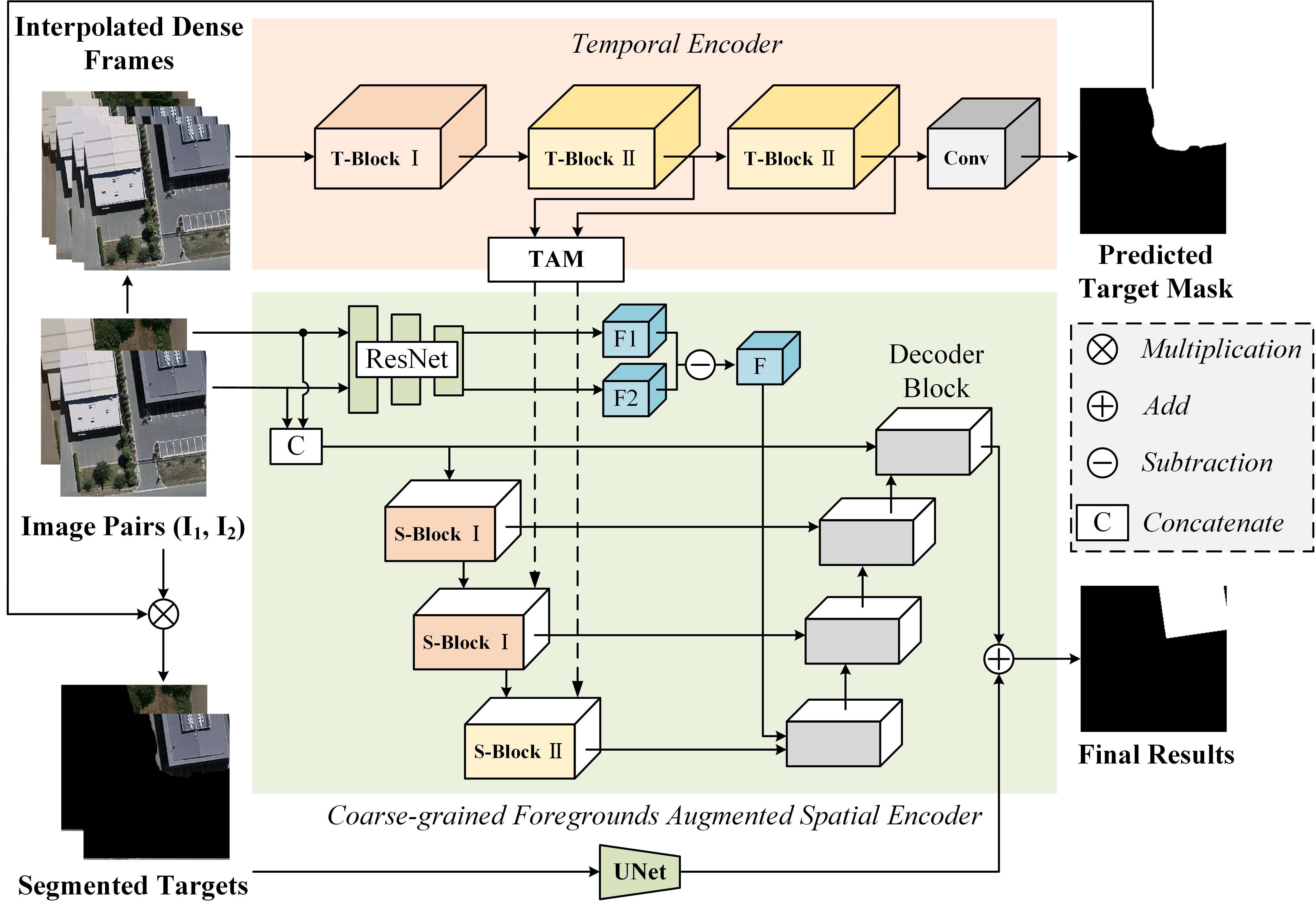} 
\caption{\textbf{Overview of Coarse-grained Temporal Mining Augmented (CTMA) framework for remote sensing image change detection.} It mainly contains two modules, i.e., Temporal Encoder (TE) and Coarse-grained Foregrounds Augmented Spatial Encoder (CFA-SE). 
Given the bi-temporal images, we first utilize TE to extract the feature representations containing temporal information and generate a preliminary mask map. 
Subsequently, we introduce CFA-SE to integrate global and local information of image pairs, and further optimize the results with a mask augmented strategy. 
This strategy dexterously leverage the initial mask map generated by TE as prior knowledge to guide CFA-SE in producing more accurate detection results. 
In addition, as a supplement to it, we also add a motion augmented strategy to consider the motion information within CFA-SE for the better overall performance.
}
\label{framework}
\end{figure*}

As illustrated in Fig.~\ref{framework}, we propose a novel Coarse-grained Temporal Mining Augmented framework for remote sensing change detection, which consists of Temporal Encoder (TE) and Coarse-grained Foregrounds Augmented Spatial Encoder (CFA-SE).
Given the bi-temporal images, we first employ Temporal Encoder (TE) to yield the feature representations containing temporal information and generate a preliminary mask.
To be concrete, we transform the input bi-temporal images into video frame sequences to enhance the extraction of temporal information. 
Subsequently, the temporal encoder processes this video data to derive the preliminary detection results based on temporal modeling, known as coarse change map.
To further improve the accuracy of detection, we introduce a novel Coarse-grained Foregrounds Augmented Spatial Encoder (CFA-SE).
This encoder not only considers the global and local information of the input image pair but also improves the extraction of motion information to produce the final probability map for remote sensing change detection.
By exploring the reliable local variation regions in the coarse change map and incorporating the threshold segmentation technique, the encoder generates an accurate mask.
In the spatial modeling phase, our approach integrates global-local branches to encode and decode complete images and images with erased background regions.
Finally, by fusing the outputs of these two branches, we obtained a more accurate change map. 
The whole process is trained end-to-end, ensuring efficient and accurate performance. 

\subsection{Temporal Encoder}
\label{sec:TE}
Given the bi-temporal images, we first refer to work~\cite{lin2022transition} to construct a pseudo-video frame sequence from the input image pair through video transformation technology to obtain a more detailed view of the temporal data. 
This method not only avoids relying on external data or manually using linear interpolation but also directly creates new frames between two known frames, ensuring a uniform distribution of these frames on the timeline. 
The pseudo-video consists of $N$ frames, where the first and last two frames are directly taken from the original images $I_1$ and $I_2$, while the $n$-th frame $X_{n}$ ($0 < n < N$) is obtained by interpolation. 
In particular, the generation of the $n$-th frame $X_{n}$ follows the following general formula: 
\begin{equation}
X_{n}=I_{1}+\frac{n}{N-1}\left ( I_{1}-I_{2} \right )  
\label{eq:equ0}
\end{equation}
This process is actually a sampling of frames in a virtual video, which accurately depicts the linear transition of all pixels in time, and the temporal resolution (or frame rate) of the video is inversely proportional to the degree of refinement of this transition. 
In addition, by processing the bi-temporal input as a subset, the constructed frame sequence effectively prevents information loss, ensuring the integrity and accuracy of data.

As shown in Fig.~\ref{framework}, we introduce a Temporal Encoder (TE), which mainly consists of a downsampling layer (T-Block I), two temporal blocks (T-Block II), a temporal aggregation module (TAM) and a convolutional layer.
T-Block I is a streamlined convolutional neural network (CNN), which uses a $3 \times 9 \times 9$ kernel size and a $1 \times 4 \times 4$ stride, facilitating a quadruple downsampling of feature maps.
This configuration allows the temporal encoder to focus on temporal information more efficiently.
On the other hand, T-Block II focuses on capturing dynamic motion information closely related to temporal changes. 
It adopts a series of 3D ResBlock modules similar to those described in~\cite {hara2018can}. 
These modules employ 3D convolutional layers with a kernel size of 3, targeting features pertaining to temporal events across different temporal dimensions. 
Its construction involves a series of $1 \times 1 \times 1$ 3D convolutional layers, each followed by a batch normalization (BN) layer~\cite{ioffe2015batch} and ReLU activation function~\cite{glorot2011deep}. 
Specifically, T-Block II includes an initial $1 \times 1 \times 1$ 3D convolutional layer followed by a duplication of two identical configurations, aiming to maintain the simplicity and computational efficiency of the network architecture. 
Additionally, T-Block II introduces a residual structure, allowing the input of the module to be connected to the output before the final ReLU activation function through a  $1 \times 1 \times 1$ 3D convolutional layer after passing through the BN layer. 
This design reduces the computational cost~\cite{he2016deep} by first reducing and then restoring the number of channels through two $1 \times 1 \times 1$ convolutional layers, thus enabling the construction of a broader and deeper network structure.

\begin{figure*} 
    \centering
    \includegraphics[width=1\linewidth]{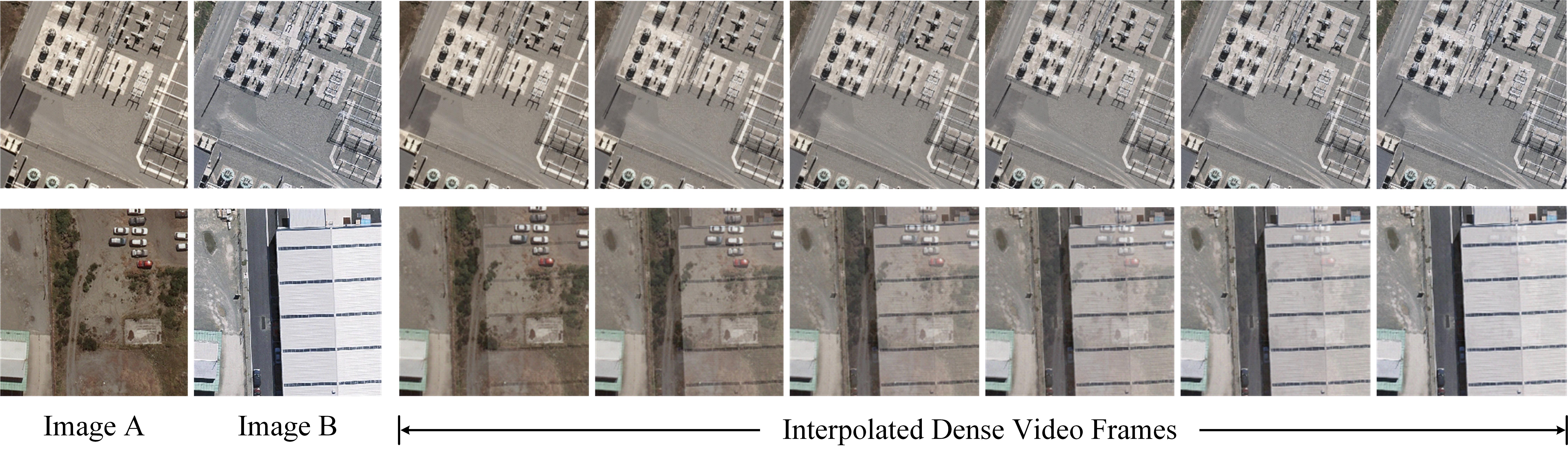}
    \caption{Qualitative results of the interpolated dense video frames on WHU-CD dataset.}
    \label{fig:inter_video}
\end{figure*}

In order to strengthen the interaction between temporal and spatial information, we introduce a temporal aggregation module (TAM) inspired by the previous work~\cite{lin2022transition}. 
Let the output $H \in \mathbb{R}^{B \times c \times T \times h \times w}$ of T-Bclock II, where $B$ symbolizes the batch size, $c$ signifies the number of channels, $T$ represents the frame rate, and $h$ and $w$ denote the height and width of the feature map, respectively.
We first perform global average pooling and global maximum pooling operations on the temporal dimension for 3D features.
Following this, the pooled features are connected in channel dimension to yield a 2D feature descriptor with $2c$ channel.
Subsequently, through a $1 \times 1$ convolutional layer with $c$ filters, combined with BN and ReLU activation function, the aggregated feature map is transformed pointwise to obtain the final feature representation $\tilde{H} \in \mathbb{R}^{B\times c \times h \times w}$.
Finally, through a convolutional layer and applying threshold segmentation, we obtain the predicted target mask matrix $M$, which can be formulated as follows,
\begin{equation}
    M=threshold(f(\tilde{H}))
\label{eq:equ1}
\end{equation}
where $M\in \mathbb{R}^{B \times 1 \times H \times W}$, $H$ and $W$ correspond to the original image dimensions.
$f(\cdot)$ denotes the a convolutional operation.

\subsection{Coarse-grained Foregrounds Augmented Spatial Encoder} 
In order to improve the accuracy and efficiency of change detection, we propose a novel Coarse-grained Foregrounds Augmented Spatial Encoder (CFA-SE) module. This module uniquely combines both global and local information and utilizes motion-augmented and mask-augmented strategies to obtain more precise change detection. The core components of the proposed CFA-SE module mainly contain bi-temporal image fusion module, motion augmented strategy, and mask augmented strategy.

Inputting the bi-temporal images, how to merge them efficiently in a change detection framework is an important and complex task. In the change detection framework, how to effectively integrate bi-temporal images is a crucial and complex task. Traditional fusion methods, such as point-to-point difference and channel-level splicing, have their own advantages, but they also have limitations. Although point-to-point difference can directly reveal the difference at pixel level, it overlooks the interactive information between images. On the other hand, channel-level splicing can merge information from two images but may neglect individual image feature extraction. To overcome these limitations, we adopt two complementary strategies in the bi-temporal image fusion module of CFA-SE.
As shown in Figure~\ref{framework}, we build two sub-networks for global image pairs.
The one branch uses ResNet~\cite{he2016deep} as the backbone network for feature extraction and captures difference features between two images through differential operation, which can be denoted as,
\begin{equation}
    F=ResNet(I_1)-ResNet(I_2)
    \label{eq:equ2}
\end{equation}
where $F \in \mathbb{R}^{B \times c' \times h' \times w'}$. 
The other branch combines the channel dimensions of the two images into a high-dimensional image pair, utilizing the Unet~\cite{ronneberger2015u} network structure to extract the change information. 
This method effectively combines pixel-level difference features and interactive information between images, enabling us to comprehensively and deeply detect changes in bi-temporal images.

To be specific, we incorporate two types of S-Block (i.e., S-Block I and S-Block II) for Unet architecture, as recommended by~\cite{lin2022transition}, in order to decrease the spatial size of the image and increase the channel size of image.
Both types of S-Blocks begin with a sequence of convolutional-BN-ReLU-convolutional-BN layers and conclude with a pooling layer, with the exception that S-Block II includes an extra convolutional layer (comprising the corresponding BN and ReLU layers). 
The use of maximum pooling for downsampling enables the capturing of multi-scale spatial context through the stacking of S-Blocks. 
Furthermore, a residual connection is established between the output of the initial ReLU layer and the output of the final normalization layer within the block. 
In order to extract change information from the spatial and temporal cues captured by the encoder, a simple yet efficient decoder is further designed, consisting of four basic decoding blocks. 
These decoding blocks predict the probability maps in an asymptotic manner by feeding the features from the previous decoding block or the underlying convolution (i.e., decoding features) with the features from the corresponding scales of the spatial encoder (i.e., encoding features). 
It is worth noting that the initial encoded feature fed into the global decoder is the output of S-Block, whereas the other input feature is sourced from the disparity features $F$ extracted by ResNet network, thus establishing the structure of Unet network.
In particular, the last decoded feature is first upsampled to fit the size of the encoded feature mapping. 
Then, the encoded and decoded features are associated by splicing them in the channel dimension, and the fused features are resolved by stacking two convolutional layers. 
To increase nonlinearity, we employ a ReLU activation function and alternate batch normalization layers to stabilize the training process.

In addition, residual concatenation is used in the second convolutional layer to further optimize the learning of features. 
Special emphasis should be placed on the fact that the original diachronic image is transferred to the final decoding block, which helps to preserve the spatial details of the image. 
Finally, the probability maps generated by the global-local decoder are fused and binarized in a weighted manner with a threshold of 0.5 to obtain the final variogram.
By integrating the two subnetworks and implementing interactive fusion operations in the decoder, CFA-SE not only improves the accuracy of change detection but also enhances the robustness of model.
In this way, although the global and local information of the bi-temporal image is considered, the importance of motion information and mask guidance is ignored. 
Therefore, in order to generate more accurate change detection results, we introduce motion augmented and mask augmented strategies.

\textbf{Motion Augmented Strategy.}
According to the detailed description in Section~\ref{sec:TE}, we successfully utilize T-block II to capture motion information during temporal changes. 
Considering that the proposed CFA-SE may overlook such crucial motion information, we implement a motion augmented strategy. Specifically, we first extract the 3D output containing motion information from T-block II, and then use temporal aggregation module (TAM) as suggested in~\cite{lin2022transition} to convert these 3D features into 2D features. 
These transformed 2D features are used as additional inputs for S-block I and S-block II in spatial encoder, thereby effectively facilitating the integration of motion information and enhancing the capability of the model to detect changes in remote sensing images.

\textbf{Mask Augmented Strategy.}
To improve the accuracy of change detection, we implement a mask augmented strategy. 
This strategy dexterously leverages the preliminary change map acquired through Temporal Encoder (TE) in section~\ref{sec:TE} as prior knowledge, delving deeper into the change information of local areas. 
Initially, we apply a customized threshold segmentation onto the coarse variable probability map output by TE, transforming it into a binary image containing only 0 and 1.
Subsequently, this binary image is upsampled to match the spatial resolution of the original image, producing an accurate mask image. 
We then utilize this mask image to element-wise multiply the original image pairs $(I_1, I_2)$. 
This operation aims to retain the regions indicated by a value of 1 in the mask image (representing potential changes) while eliminating the areas masked with a value of 0 (indicating non-changing regions). 
This process yields the segmented object image pair, which serves as input.
The object image pair is then concatenated and fed into Unet~\cite{ronneberger2015u} network, enabling the network to focus on learning the change regions within images.
Finally, we fuse the probability map output from Unet with the probability map generated by the global-local decoder to generate the final probability map.
This strategy effectively eliminates the interference of static background in bi-temporal image pair, enabling the model to focus more on detecting changing regions and significantly enhancing the accuracy of detection results.

\subsection{Loss Function} 

To improve the performance of model during the training process, we add the weighted Binary Cross-Entropy (BCE) loss into the output probability maps of the TE and CFA-SE, respectively. 
This addition serves as a means of supervision and constraint.
Given that the two probability maps are labeled as $P_1$ and $P_2$ respectively, the total loss can be expressed as:
\begin{align}
& \mathcal{L}_{total} = \alpha \mathcal{L}_1(P_1, Y) + (1-\alpha) \mathcal{L}_2(P_2, Y)
\label{equ:eq3}
\end{align}
where $\mathcal{L}_1$ and $\mathcal{L}_2$ represent the weighted binary cross-entropy loss functions.
More detailed information can be found in~\cite{zhang2020deepemd}.
$Y$ denotes the ground-truth change map.
The parameter $\alpha$ is a balancing hyper-parameter, which is empirically set to 0.5 in all of our experiments.

\section{Experiments} \label{experiments}
\subsection{Dataset Description}   
In our experimental study, we use three well-known remote sensing image datasets for remote sense change detection, i.e., \textbf{SVCD}~\cite{lebedev2018change}, \textbf{LEVIR-CD}~\cite{chen2020spatial}, and \textbf{WHU-CD}~\cite{ji2018fully}.
Each dataset is briefly described as follows:

$\bullet$ \textbf{SVCD}~\cite{lebedev2018change} is a comprehensive change detection dataset consisting of 11 pairs of accurately aligned remote sensing images obtained from Google Earth. 
The images exhibit a diverse spatial resolution ranging from 3 cm to 100 cm per pixel.
Through careful random cropping and rotation procedures, we generate 16,000 pairs of $256 \times 256$ image fragments. 
These fragments are thoughtfully allocated, with 10,000 pairs designated for training, 3,000 pairs for validation, and 3,000 pairs for testing. 
Each pair of image fragments contains at least one altered pixel, ensuring the richness of changing information in the dataset. 
It is worth noting that SVCD employs a very strict definition of data change, and only substantial modifications in the semantic class of an object are deemed as changes based on the fundamental annotation. 
This means that subtle differences such as seasonal natural changes in leaves are not reflected in the ground reality label. 
This feature puts forward higher requirements for change detection methods, requiring algorithms to achieve higher levels in radiation calibration accuracy and semantic information utilization.



$\bullet$ \textbf{LEVIR-CD}~\cite{chen2020spatial} is specifically designed for detecting changes in buildings, consisting of 637 pairs of remote sensing image tiles, each with a high-resolution size of $1024 \times 1024$ pixels. 
These images possess a detailed spatial resolution of 0.5 meters per pixel, enabling accurate delineation of subtle building alterations.
The dataset encompasses a total of 31,333 different change instances, covering a diverse range of building categories, thus ensuring comprehensive dataset breadth and representation.
To maintain consistency and accuracy of training and evaluation, LEVI-CD strictly follows the predefined training/validation/test data split scheme by the dataset authors. 
During data processing, a sliding window technique, as described in~\cite{lin2022transition}, is employed to extract $256 \times 256$ pixel chips from the original image tiles. 
This approach significantly enhances the ease of training and evaluating models on large raster images. 
During the training phase, in order to increase the diversity of samples and improve the generalization ability of the model, the chips are overlapped at 128-pixel strides, resulting in more training samples. 
Subsequently, during the verification and test phase, in order to ensure the objectivity of the evaluation results, the verification and test chips are extracted in a non-overlapping manner, with a step size configured to 256 pixels. 
This strategy not only guarantees the effectiveness of model training, but also comprehensively addresses the inherent complexities associated with processing large-scale raster images.


$\bullet$ \textbf{WHU-CD}~\cite{ji2018fully} dataset consists of two high-resolution aerial images, each with dimensions of up to $32507 \times 15354$ pixels and an accurate spatial resolution of 0.3 meters per pixel. 
Compared with the LEVIR-CD~\cite{chen2020spatial} dataset, WHU-CD dataset focuses on the detection of building changes. 
Our research method begins by carefully cropping the original image into non-overlapping $256 \times 256$ pixel image chips. 
Subsequently, these chips are randomly divided into three distinct subsets: 6,096 samples for training, 762 samples for validation, and another 762 samples for testing. 
The data partitioning strategy employed in this dataset is consistent with the method described in the previous work~\cite{chen2021remote}, which ensures the integrity and reliability of the experimental results.


\subsection{Evaluation Metric}
In our experimental evaluation, we use three key criteria to comprehensively evaluate the performance of change detection algorithms, namely Precision, Recall, and F1 scores. 
Each metric plays a different role in the evaluation process and aims to reveal the performance of the algorithm from a different perspective.
Precision focuses on the proportion of true positive samples among those predicted by the algorithm, which reflects the accuracy of the algorithm. 
We can define it as,
\begin{align}
& P = \frac{N_{tp}}{N_{tp} + N_{fp}}
\end{align}
where $N_{tp}$ and $N_{fp}$ represent the number of true positive and false positive samples, respectively. 
Recall measures the ability of the algorithm to correctly identify all true positive samples, indicating the completeness of the algorithm. 
It can be denoted as follow,
\begin{align}
& R = \frac{N_{tp}}{N_{tp} + N_{fn}}
\end{align}
where $N_{fn}$ represents the number of false negative samples. 
The F1 score is the harmonic average of Precision and Recall, providing us with a comprehensive evaluation metric that balances accuracy and completeness. 
It can be formulated as,
\begin{align}
& F1 = 2 * \frac {P * R} {P + R}
\end{align}
In a word, these metrics provide a comprehensive overview of the performance of the change detection algorithm, ensuring the objectivity and comprehensiveness of our evaluation.


\subsection{Implementation Details} 
This section elaborates on the building blocks in the network. 
The Temporal Encoder (TE) is constructed from three T-Blocks without shared parameters. 
The local branch in Coarse-grained Foregrounds Augmented Spatial Encoder (CFA-SE) consists of three independent S-Blocks, which do not share parameters.
In the S-Block, the output channels of each Block is set to 32, 64, and 128. 
The first two S-Blocks are categorized as S-Block I, while the last one is classified as S-Block II. 
Given that video frames provide a more abundant data source compared to bi-temporal image pairs, the TE module include additional layers and convolutional filters.
Consequently, the output channels of T-Block I is set to 64, while the terminal convolutional layers of the two T-Blocks (i.e., T-Block II) utilize 256, 256, 512, and 512 filters, respectively.
A fixed threshold of 0.5 is set for segmentation when constructing the mask.

Three datasets, i.e., SVCD~\cite{lebedev2018change}, LEVIR-CD~\cite{chen2020spatial} and WHU-CD~\cite{ji2018fully}, are used in the experiment. 
For the LEVIR-CD dataset, we follow the processing method of the benchmark work~\cite{lin2022transition}, which extracts $256 \times 256$ image blocks from the original image by sliding the window. 
In order to ensure the sufficiency of the sample, the training set image blocks are overlapped with a step size of 128 pixels, while the validation set remains non-overlapped with a step size of 256 pixels. 
In the testing phase, we also use 256 pixel steps without overlapping, and the final prediction score is determined by averaging the prediction scores of all overlapping inference windows for a given pixel.

All models are implemented based on the PyTorch framework and trained on an Intel Xeon Silver 4314 CPU and a single NVIDIA GeForce RTX 3090 GPU. 
On the three datasets, we uniformly set the batch size to 8 and use the Adam optimizer to update the network parameters.
During the linear interpolation of video frames, we build 8 frames and set the weight coefficient of auxiliary loss to 0.4.
The balance weight of binary cross-entropy loss is fixed at 0.5 for both TE and CFA-SE.
To enhance data diversity, training data is randomly flipped, moved, and rotated 90-degrees, followed by training in small batches. 
We use different learning rates and iterations for different datasets. 
Specifically, for the SVCD dataset, the initial learning rate is set to 0.0004 and reaches convergence in 260,000 iterations. 
For the LEVIR-CD dataset, the initial learning rate is set to 0.002 and the number of iterations is 220,000. 
For the WHU-CD dataset, the training starts with a learning rate of 0.0004 and concludes after 160,000 iterations. 
We employ a step decay strategy to adjust the learning rate, with attenuation rates and step sizes set to 0.1 and 70 for SVCD dataset, 0.2 and 60 for LEVIR-CD dataset, and 0.2 and 105 for WHU-CD dataset, respectively.
In addition, we also set different weight factors for the output probability map of CFA-SE according to the characteristics of the dataset. 
Following each training round, a validation step is undertaken to identify the model with the highest F1 score as the best model to evaluate its performance on the test subset.

\begin{table*}[t]
\caption{Comparison results with other state-of-the-art (SOTA) models on three remote sensing change detection datasets. All scores are reported in percentage (\%). \textbf{Bold} and \underline{underline} denote the first and second experimental results, respectively.} 
\label{benchmarkResults}
\centering
\begin{tabular}{lccccccccccc}
\toprule
\textbf{Method} & \multicolumn{3}{c}{\textbf{SVCD}} && \multicolumn{3}{c}{\textbf{LEVIR-CD}} && \multicolumn{3}{c}{\textbf{WHU-CD}}\\
\cmidrule{2-4} \cmidrule{6-8} \cmidrule{10-12}
&P & R & F1 && P & R & F1  && P & R & F1  \\
\toprule
FC-EF~\cite{daudt2018fully}   & 87.41 & 51.80 & 65.05 && 90.64 & 78.84 & 84.33 && 80.65 & 78.86 & 79.74  \\
FC-Siam-Conc~\cite{daudt2018fully}  & 92.49 & 58.78 & 71.88 && 92.55 & 84.40 & 88.28 && 70.35 & 87.64 & 78.05  \\
FC-Siam-Di~\cite{daudt2018fully}   & 93.65 & 54.32 & 68.76 && 91.21 & 81.18 & 85.90 && 67.88 & 83.59 & 74.92  \\
CDNet~\cite{alcantarilla2018street}   & 92.51 & 87.77 & 90.07 && 91.52 & 88.05 & 89.75 && 92.49 & 88.09 & 90.23  \\
STANet~\cite{chen2020spatial}   & 95.17 & 92.88 & 94.01 && \textbf{93.38} & 86.58 & 89.85 && 92.69 & 88.99 & 90.80  \\
BIT~\cite{chen2021remote}  & 96.07 & 93.49 & 94.76 && 90.80 & 89.74 & 90.27 && 84.91 & 87.20 & 86.04  \\
L-UNet~\cite{papadomanolaki2021deep}   & 96.52 & 94.41 & 95.45 && 93.18 & 88.64 & 90.85 && 79.00 & \underline{89.38} & 83.87  \\
DSIFN~\cite{zhang2020deeply}  & 97.65 & 94.85 & 96.23 && 92.45 & 87.06 & 89.67 && \textbf{96.26} & 86.77 & 91.27  \\
SNUNet~\cite{fang2021snunet}   & 98.09 & 97.42 & 97.75 && 93.08 & \underline{89.90} & \textbf{91.47} && 89.90 & 86.82 & 88.33  \\
P2V-CD~\cite{lin2022transition} & \underline{98.55} & \underline{98.03} & \underline{98.29} && \underline{93.20} & 89.73 & 91.14 && 95.09 & 88.69 & \underline{91.78}  \\
OURS     & \textbf{98.67} & \textbf{98.45} & \textbf{98.56} && 92.86 & \textbf{90.26} & \underline{91.15} && \underline{95.32} & \textbf{91.99} & \textbf{93.62}  \\
\bottomrule
\end{tabular}
\end{table*}

\subsection{Comparison with State-of-the-Art Methods}  
The method was comprehensively validated on three benchmark datasets and compared with 10 other state-of-the-art change detection models, including FC-EF~\cite{daudt2018fully}, FC-Siam-Conc~\cite{daudt2018fully}, FC-Siam-Di~\cite{daudt2018fully}, CDNet~\cite{alcantarilla2018street}, STANet~\cite{chen2020spatial}, BIT~\cite{chen2021remote}, L-UNet~\cite{papadomanolaki2021deep}, DSIFN~\cite{zhang2020deeply}, SNUNet~\cite{fang2021snunet}, and P2V-CD~\cite{lin2022transition}. 
The quantitative evaluation results for the three datasets are detailed in Table~\ref{benchmarkResults}, with all data units represented in percentages.

\textbf{Results on SVCD dataset.} 
On SVCD dataset~\cite{lebedev2018change}, our proposed method demonstrates significant improvements over the L-Net method~\cite{papadomanolaki2021deep}, which is based on a UNet-like architecture, with gains of 2.15\%, 4.04\% and 3.11\% in accuracy, recall and F1 score metric, respectively.
Compared with DSIFN method~\cite{zhang2020deeply} that focuses on high-resolution bi-temporal remote sensing images, our proposed method shows enhancements of 1.02\%, 3.60\% and 2.33\% in accuracy, recall and F1 score metric, respectively. 
In addition, compared with P2V-CD method~\cite{lin2022transition} which also employs a pair-to-video transformation strategy, our proposed method achieves improvements of 0.12\%, 0.42\% and 0.27\% in accuracy, recall and F1 score metric, respectively.
These results show that the proposed CTMA method has better performance on this dataset.

\textbf{Results on LEVIR-CD dataset.} 
On the LEVIR-CD dataset~\cite{chen2020spatial}, our proposed method exhibits improvements in accuracy (0.41\%), in recall (3.2\%), and the F1 score metric (1.48\%) compared to the DSIFN method~\cite{zhang2020deeply}. 
When compared with the P2V-CD method~\cite{lin2022transition}, although the accuracy of our proposed method is slightly lower, it achieves a noteworthy 0.53\% enhancement in recall metric. 
This result could be attributed to the diverse architectural variations and intricate environmental contexts present in Levi-CD dataset. 
In the process of pursuing high recall metric, our proposed method adopts a relatively loose strategy for predicting changing regions, leading to the introduction of some false positive examples that affect the accuracy to some extent. 
Nevertheless, it should be noted that the F1 score metric is a harmonic average of the accuracy and recall metric, which enables a more comprehensive evaluation of the model's performance. 
Although slightly less accurate than the P2V-CD method, our method still performs better on the F1 score metric, which fully demonstrates the overall performance advantage of the proposed CTMA method.

\textbf{Results on WHU-CD dataset.} 
On the WHU-CD dataset~\cite{ji2018fully}, our proposed method showcases competitive performance. 
For example, compared with the L-UNet method, our proposed method achieves a significant improvements, including a 16.32\% increase in accuracy metric, a 2.61\% increase in recall metric, and a significant improvement of 9.75\% in F1 score metric.
Furthermore, compared with SNUNet method based on channel attention mechanism, our proposed method exhibits exceptional strength, delivering outstanding results with improvements of 5.42\%, 5.17\% and 5.29\% in accuracy, recall and F1 score metric, respectively. 
Compared with the P2V-CD method, our proposed method also performs well.
We achieve a 0.23\% enhancement in accuracy metric, a 3.3\% increase in recall metric, and a 1.84\% rise in F1 score metric.
Similarly, our proposed method demonstrates competitive performance compared to most other SOTA methods. 
These results fully validate the effectiveness and superiority of the proposed CTMA method.


\subsection{Ablation Study} 
In this section, we provide a detailed analysis of the different components, hyper-parameter settings, different mask thresholds and different interpolated video frames in the proposed CTMA method on the WHU-CD dataset~\cite{ji2018fully}. 
By conducting a detailed analysis of the effectiveness of these pivotal factors, we aim to elucidate their impact on model performance and provide readers with a more comprehensive understanding and reference.


\begin{table*}[!t]
\caption{Ablation study of different components in the proposed CTMA model on WHU-CD dataset.
All these scores are written in percentage (\%).
\textbf{Bold} denotes the best result.
} 
\label{tab:DifferentComponentsAnalysis}  
\centering 
\begin{tabular}{c|c|cc|c|cccc}
\toprule
\multirow{2}{*}{\#} & \multirow{2}{*}{TE}   & \multicolumn{2}{c|}{CFA-SE}  & \multirow{2}{*}{MA}         & \multirow{2}{*}{P} & \multirow{2}{*}{R} & \multirow{2}{*}{F1} & \multirow{2}{*}{OA} \\ 
    &              & SE       & ResNet         &           &           &            &          &      
\\     \hline 
1   &\textcolor{SeaGreen4}{\cmark} &\textcolor{SeaGreen4}{\cmark} &\textcolor{DarkRed}{\xmark} &\textcolor{DarkRed}{\xmark}                            &95.09 &88.69 &91.78 &99.23 \\
2   &\textcolor{SeaGreen4}{\cmark} &\textcolor{SeaGreen4}{\cmark} &\textcolor{DarkRed}{\xmark} &\textcolor{SeaGreen4}{\cmark}                            &95.32 &88.80 &91.94 &99.25       \\
3   &\textcolor{SeaGreen4}{\cmark} &\textcolor{SeaGreen4}{\cmark} &\textcolor{SeaGreen4}{\cmark} &\textcolor{DarkRed}{\xmark}                            &\textbf{96.19} &90.13 &93.06 &99.35      \\
4   &\textcolor{SeaGreen4}{\cmark} &\textcolor{SeaGreen4}{\cmark} &\textcolor{SeaGreen4}{\cmark} &\textcolor{SeaGreen4}{\cmark}                            &95.32 &\textbf{91.99} &\textbf{93.62} &\textbf{99.40}     \\
\bottomrule
\end{tabular}
\end{table*}

\textbf{Analysis of different components.}
In order to comprehensively evaluate the contribution of each individual component of the proposed CTMA method on network performance, we conduct an ablation experiment for component analysis on the WHU-CD dataset. 
The proposed CTMA method mainly combines temporal encoder (TE), spatial encoder (SE), ResNet, and mask augmented (MA) strategy. 
The experimental findings are visually depicted in Table~\ref{tab:DifferentComponentsAnalysis}.
Specifically, `TE' captures and learns the dynamic motion information between dense frames after frame insertion, which is the key to understanding temporal changes. 
`SE' is the basic element of the CFA-SE module, which focuses on extracting features sensitive to spatial details from spatial dimensions. 
Meanwhile, `MA' can enhance the feature representation by introducing mask to further improve the model performance. 
The following conclusions can be drawn from the experimental results: 
1) Compared \#1 with \#2, it can be clearly observed that the introduction of MA improves the network performance, which proves the effectiveness of MA strategy. 
2) Compared \#1 with \#3, it reveals that integrating ResNet in CFA-SE module not only enhances the network's ability to learn diverse and multi-level information, but also enhances its ability to decipher complex scenarios. 
3) The experimental results of \#4 show that the CTMA method achieves an optimal level of performance when all components work together. 
This fully demonstrates the complementary nature of different components, collectively establishing an effective and comprehensive change detection framework.
In summary, the results of ablation experiments validate the rationale and necessity behind the design of each component in the proposed CTMA method.
Their collective integration improves the overall performance of the network in the change detection task.

\begin{table}[t]
\caption{Analysis results of hyper-parameter on WHU-CD dataset.
All these scores are written in percentage (\%).
\textbf{Bold} denotes the best result.}  
\label{tab:hyperparameter}  
\small 
\centering 
\begin{tabular}{c|cccc}
\toprule
Values     &P & R & F1 & OA \\
\hline
0.1   &95.09&91.64&93.34&99.37 \\ 
0.2   &96.06&89.82&92.84&99.33 \\ 
0.3   &95.32&\textbf{91.99}&\textbf{93.62}&\textbf{99.40} \\ 
0.4   &\textbf{95.95}&90.37&93.08&99.35 \\
0.5   &95.37&91.08&93.18&99.36 \\
\bottomrule
\end{tabular} 
\end{table}

\begin{table}[t]
\caption{Analysis results of mask threshold on WHU-CD dataset.
All these scores are written in percentage (\%).
\textbf{Bold} denotes the best result.}  
\label{tab:mask_thred}  
\small 
\centering 
\begin{tabular}{c|cccc}
\toprule
Values     &P & R & F1 & OA \\
\hline
0.4   &\textbf{95.52}&90.74&93.07&99.35 \\ 
0.5   &95.32&\textbf{91.99}&\textbf{93.62}&\textbf{99.40} \\ 
0.6   &94.55&89.86&92.15&99.26 \\ 
\bottomrule
\end{tabular} 
\end{table}

\begin{table}[t]
\caption{Analysis results of frame number on WHU-CD dataset.
All these scores are written in percentage (\%).
\textbf{Bold} denotes the best result.}  
\label{tab:frame_num}  
\small 
\centering 
\begin{tabular}{c|cccc}
\toprule
Values     &P & R & F1 & OA \\
\hline 
7   &94.88&91.5&93.16&99.35 \\ 
8   &95.32&\textbf{91.99}&\textbf{93.62}&\textbf{99.40} \\ 
9   &\textbf{95.56}&90.48&92.95&99.34 \\ 
\bottomrule
\end{tabular} 
\end{table}

\textbf{Analysis of hyper-parameter settings.}
In the CFA-SE module of the proposed CTMA method, we introduce an important hyper-parameter $\lambda$, which controls the contribution of mask augmented branch. 
In order to further explore the effect of $\lambda$ on model efficiency, we conduct a series of systematic analysis experiments on the WHU-CD dataset.
The experimental results are outlined in Table~\ref{tab:hyperparameter}. 
The experimental findings clearly indicate that as the value of $\lambda$ gradually increases, the performance of the model initially shows a steady upward trend until $\lambda=0.3$ obtained the optimal performance.
This observation shows that moderately increasing the weight of the mask augmented branch can effectively enhance the performance of the model in the remote sensing change detection task. 
However, when $\lambda>0.3$, we observe that the performance of the model begins to gradually decline. 
The reason may be as follows:
Firstly, an excessively high value may lead to the model overly relying on information from mask augmented branch, while ignoring the contributions of other important features or branches.
Secondly, information fusion between different branches needs to achieve $\lambda$ delicate balance.
An excessively high value may disrupt this balance, making it difficult for the model to effectively integrate information from different branches and affecting the final detection results.
Therefore, we default to setting $\lambda=0.3$ on the WHU-CD dataset unless otherwise specified.

\textbf{Analysis of different mask thresholds.~} 
In investigating the effect of mask threshold on model performance, we adjust the mask degree of change detection output by the temporal encoder branch to optimize the mask augmented process. 
As illustrated in Table~\ref{tab:mask_thred}, we find that the model obtains the best performance on multiple evaluation metrics when the mask threshold is set to 0.5. 
This may be because a low mask threshold may contain excessive information about the unchanged region, while a high mask threshold may disregard important details in the change region. 
Therefore, unless stated otherwise, we use a default mask threshold of 0.5 in all experiments.

\textbf{Analysis of different interpolated video frames.~}  
After inputting the initial and final images, we use interpolation techniques to generate a dense sequence of video frames. 
This process involves reasonable inference and filling of motion information between two images. 
The frame number serves as an important parameter in this process, which determines the density and smoothness of video frames. 
A higher frame number can more accurately restore the motion trajectory of objects and reduce motion blur, but it may also reduce the clarity and detail expression of each frame image due to bandwidth limitations.
On the contrary, a lower frame number alleviates the computational burden and storage requirements but may result in insufficient motion detail capture and consequent motion blur.
To investigate the effect of different frame numbers on the model performance, we refer to the previous work~\cite{lin2022transition} and conduct corresponding experimental analysis. 
The experimental findings presented in Table~\ref{tab:frame_num}, demonstrate that our model achieves the best performance with 8 frames.
Therefore, in all experiments, we default to setting 8 frames as the frame number to strike a balance between video quality and computational efficiency.

\begin{figure*}[!tbp]
    \centering
    \includegraphics[width=1\linewidth]{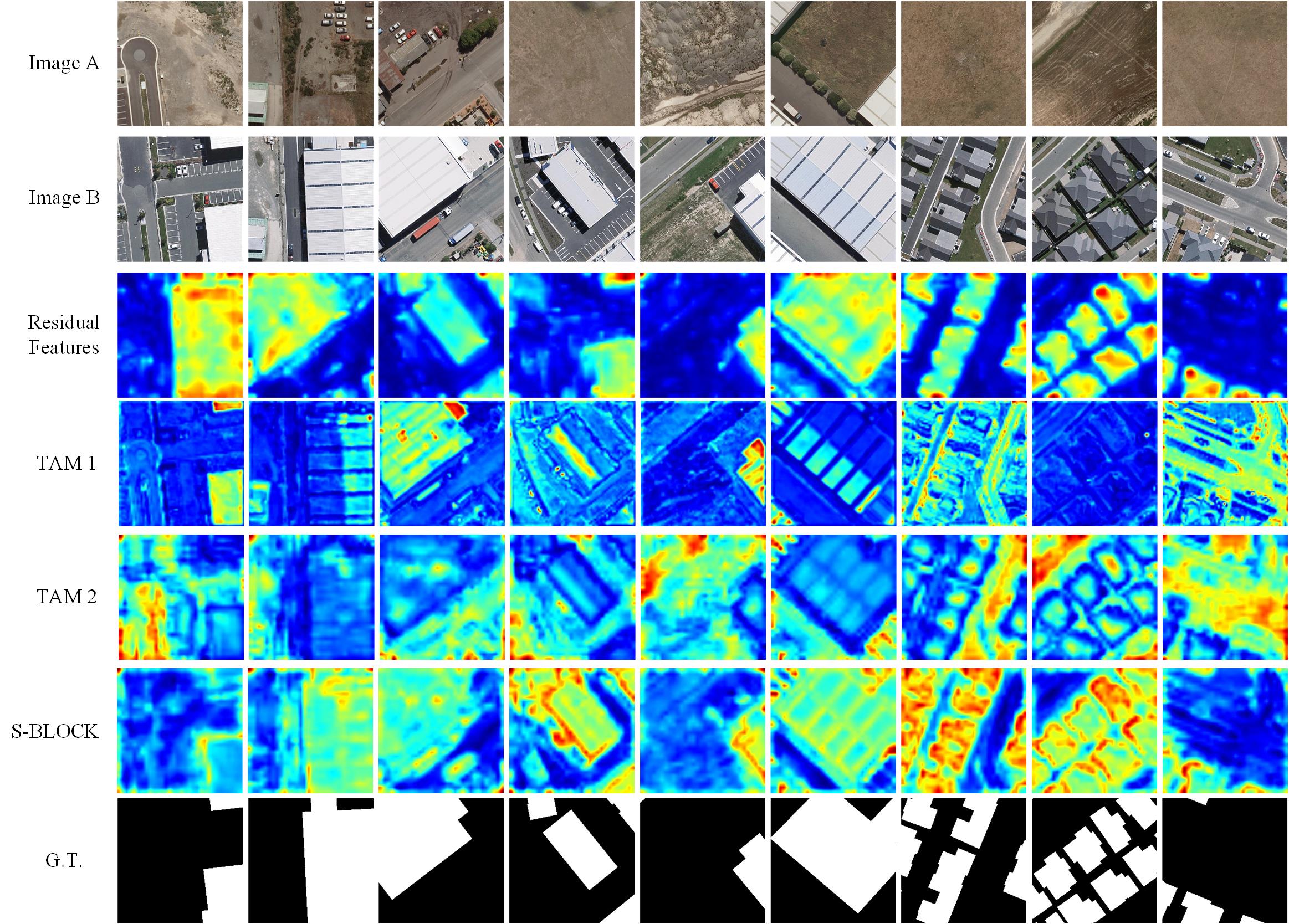}
    \caption{Visualization of feature maps learned by WHU-CD test set. The brighter the color, the greater its response value. `G.T.' denotes the ground-truth label of the corresponding image. }
    \label{fig:vis_featuremap}
\end{figure*}

\subsection{Visualization} 

\begin{figure*}
    \centering
    \includegraphics[width=1\linewidth]{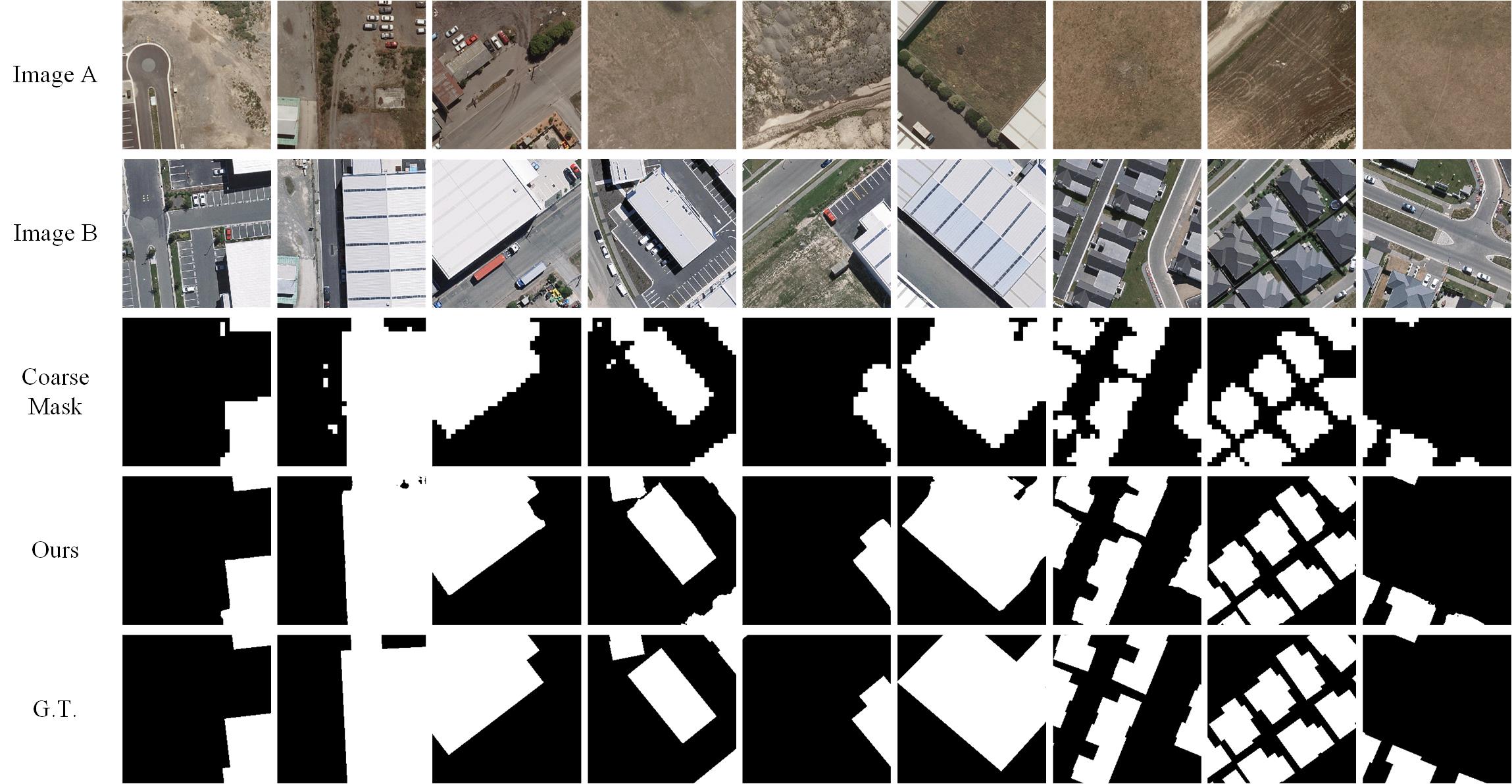}
    \caption{Visualization of course mask and change detection results acquired by the temporal encoder (TE) and the proposed CTMA method, respectively.}
    \label{fig:CD_course}
\end{figure*}

\textbf{Dense Video Frames.}
As shown in Fig.~\ref{fig:inter_video}, image A is the RGB image of the first date and image B is the RGB image of the second date. Following~\cite{papadomanolaki2021deep,lin2022transition}, we use an advanced interpolation algorithm to generate a high-density frame sequence between the two without relying on any manual labeling or external data. 
This method is not only efficient, but also can accurately model the progressive process of regional change between image A and image B. Specifically, by capturing and simulating subtle differences in pixel intensity, color distribution, and spatial structure over time, we successfully reproduce the dynamic evolution of the scene from date one to date two, providing rich and continuous data support for further analysis of applications such as land use change, environmental monitoring, or video enhancement.

\textbf{Feature Maps.} 
In order to comprehensively validate the performance of the fundamental components within our proposed CTMA method, we conduct a feature map visualization analysis on the WHU-CD dataset. 
As depicted in Fig.~\ref{fig:vis_featuremap}, for input images A and B, we successfully extract and present the visualization results of the residual features, the feature map learned by the two TAM and the S-block in the CFA-SE module respectively.
By observing these visualizations, we can find that while each module focuses on the information from different perspectives, they all exhibit a notable inclination towards the changing areas within the images, concurrently suppressing the feature expression of the non-changing areas. 
This discovery not only intuitively validates the independent value of each core component in the proposed CTMA method and the rationality of its design, but also reveals how these components work closely together to capture and enhance key change information in the image.

\textbf{Change Detection Results.}
In order to verify the effectiveness of mask augmented strategy and the CTMA model, we conduct a visual experiment of change detection results on the WHU-CD dataset, as illustrated in Fig.~\ref{fig:CD_course}.
It can be clearly observed from Fig.~\ref{fig:CD_course} that we first obtain a relatively accurate initial rough mask by use the temporal encoder. 
Subsequently, this preliminary mask serves as a robust guide to facilitate subsequent processing.
Finally, we extract more accurate change detection results.
This significant improvement not only intuitively demonstrates the potential of our model to improve the accuracy of change detection, but also strongly demonstrates the effectiveness of mask augmented strategy in combination with the CTMA model.

\begin{figure}[!tbp]
\centering
\includegraphics[width=0.48\textwidth]{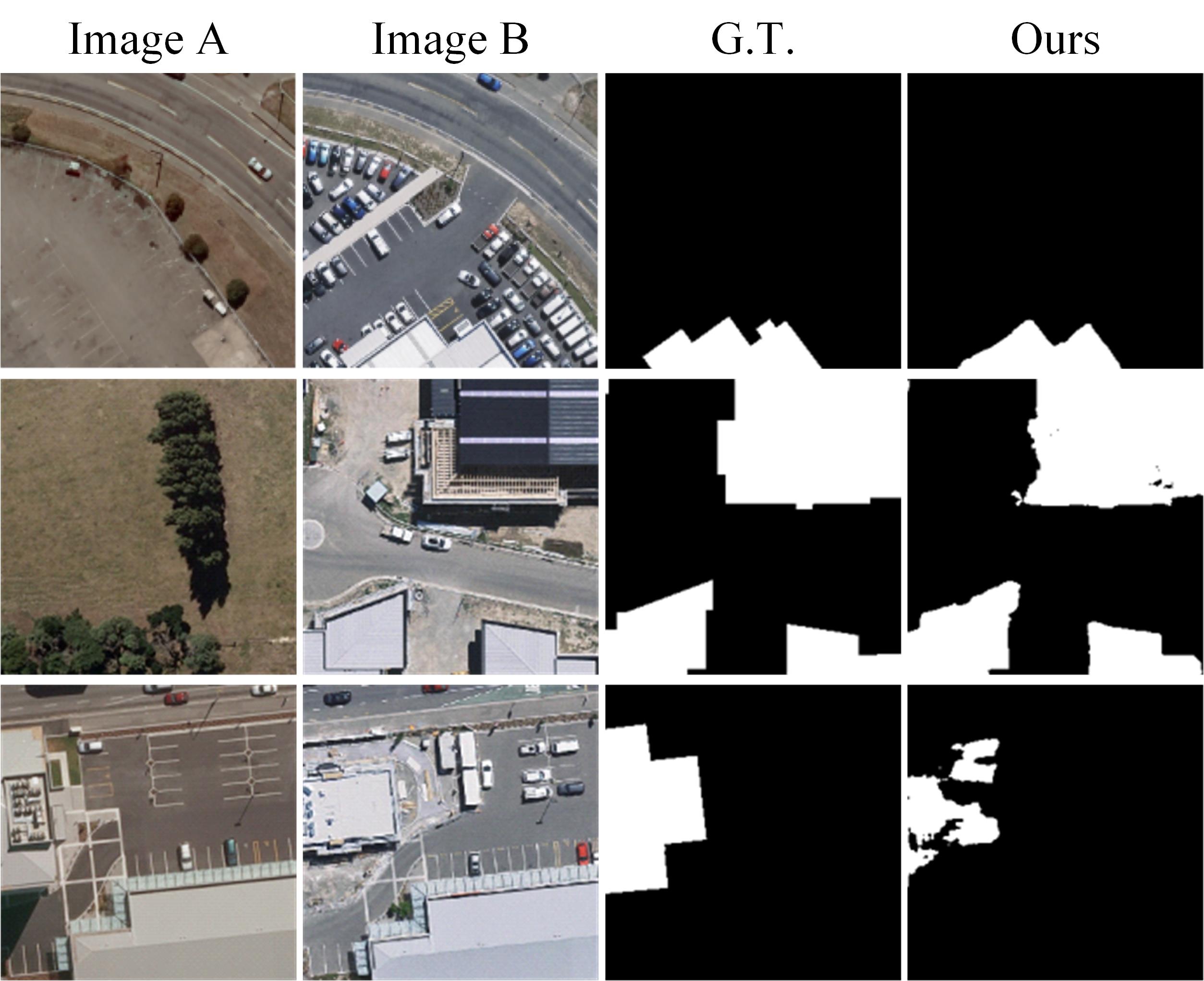}
\caption{Poor results obtained by the proposed CTMA model. `G.T.' denotes the ground-truth label of the corresponding example.} 
\label{fig:Limitation}
\end{figure}

\subsection{Limitation Analysis} 
Although the proposed CTMA model has achieved good performance on remote sensing change detection datasets, it still faces several challenges. 
As illustrated in Fig.~\ref{fig:Limitation}, the model struggles with detection effectiveness in scenarios where there are either too many or too few regions of change. For instance, the boundary detection is not clear enough (second row), and some change regions are missing (first and third rows). 

Upon conducting a thorough analysis of these shortcomings, we believe that the main possible reasons are as follows: 1) There may be errors in the temporal encoder when predicting change masks, making it inherently difficult to achieve optimal performance based on these inaccurate masks, which guide the overall learning process of the change detection framework. 
2) Although the model attempts to integrate global and local information, it lacks an effective supervision mechanism to ensure the successful integration and complementarity of these two types of information. 
This can result in incomplete or missing information in the predicted change regions. 
Therefore, we can conduct further exploration around these two aspects in the future.

\section{Conclusion} \label{conclusion}
This work builds upon the existing bi-temporal images-based framework for remote sensing change detection by introducing a novel Coarse-grained Temporal Mining Augmented (CTMA) framework. Our approach, which incorporates motion cues through the transformation of bi-temporal images into a video and the subsequent extraction of motion features, demonstrates the significance of temporal information in change detection tasks. By integrating these motion features with spatial features and utilizing a ResNet for fine-grained feature learning, our method achieves enhanced performance in predicting changed regions. The segmentation and integration of coarse-grained changed regions into decoder blocks further refine the change prediction process. The extensive experimental results across various benchmark datasets conclusively validate the effectiveness of our proposed framework. The source code for this study is made available allowing for further research and application development in the field of remote sensing image change detection.

In our future works, we will further consider adopting more lightweight and hardware-friendly networks to build our framework, such as Mamba~\cite{gu2023mamba, wang2024SSMSurvey} or RWKV~\cite{peng2023rwkv}. It will achieve a better trade-off between the detection performance and computational cost. Also, we will consider introducing semantic information perception modules to mining the locally changed regions well. We believe this will fill the holes in our final prediction as illustrated in Fig.~\ref{fig:Limitation}.


{\scriptsize
\bibliographystyle{unsrt}
\bibliography{reference}
}

\end{document}